\documentclass[11pt]{article}

\usepackage[final]{acl}

\usepackage{times}
\usepackage{latexsym}

\usepackage[T1]{fontenc}

\usepackage[utf8]{inputenc}

\usepackage{microtype}

\usepackage{inconsolata}

\usepackage{graphicx}

\usepackage{multirow}       
\usepackage{booktabs}       
\usepackage{colortbl}       
\usepackage{amsmath}        
\usepackage{amssymb}
\usepackage{makecell}       
\usepackage{adjustbox}      
\usepackage{tabularx}

\usepackage{subcaption}

\usepackage{tcolorbox} 
\definecolor{Gray}{gray}{0.9}
\definecolor{LightCyan}{rgb}{0.88,1,1}

\title{Sentinel: Decoding Context Utilization via Attention Probing for Efficient LLM Context Compression}

\author{
  \textbf{Yong Zhang}\textsuperscript{1}, 
    \textbf{Heng Li}\textsuperscript{1,2}, 
  \textbf{Yanwen Huang}\textsuperscript{1,3}, 
  \textbf{Ning Cheng}\textsuperscript{1,\textasteriskcentered},
  \\
    \textbf{Yang Guo}\textsuperscript{1}, 
  \textbf{Yun Zhu}\textsuperscript{1}, 
  \textbf{Yanmeng Wang}\textsuperscript{1}, 
  \textbf{Shaojun Wang}\textsuperscript{1}, 
  \textbf{Jing Xiao}\textsuperscript{1}, 
 \\
  \textsuperscript{1} Ping An Technology (Shenzhen) Co., Ltd., China\\
    \textsuperscript{2}   University of Science and Technology of China \\
  \textsuperscript{3} University of Electronic Science and Technology of China\\
    \texttt{zhangyong.chuck@gmail.com}
}

\begin{document}
    \maketitle

\begin{abstract}
Retrieved contexts are often long, noisy, and redundant, motivating query-aware compression before answer generation. Native attention provides signals for context selection, but heterogeneous head behaviors complicate their direct aggregation into sentence-utility scores. We introduce Sentinel, a sentence-level compressor that learns to combine these signals through a signed linear readout across layers and heads. Only the readout is trained using existing QA evidence annotations, while the backbone remains frozen and sentence scoring requires no autoregressive generation. Across twelve LongBench tasks, Sentinel with a 0.5B backbone outperforms Full Context on multiple tasks under a 2,000-token budget with each of three readers: Qwen2.5-7B-Instruct, Llama-3.1-8B-Instruct, and DeepSeek V4.1 Flash. Despite English-only QA supervision, the same readout transfers to Chinese tasks without task-specific retraining. Length-stratified profiling further shows lower compression latency and GPU memory usage than LLMLingua-2 and CPC.\footnote{Our code is available at \url{https://github.com/yzhangchuck/Sentinel}.}
\end{abstract}

\section{Introduction}

\begin{figure*}[!tp]
    \centering
    \includegraphics[width=\linewidth]{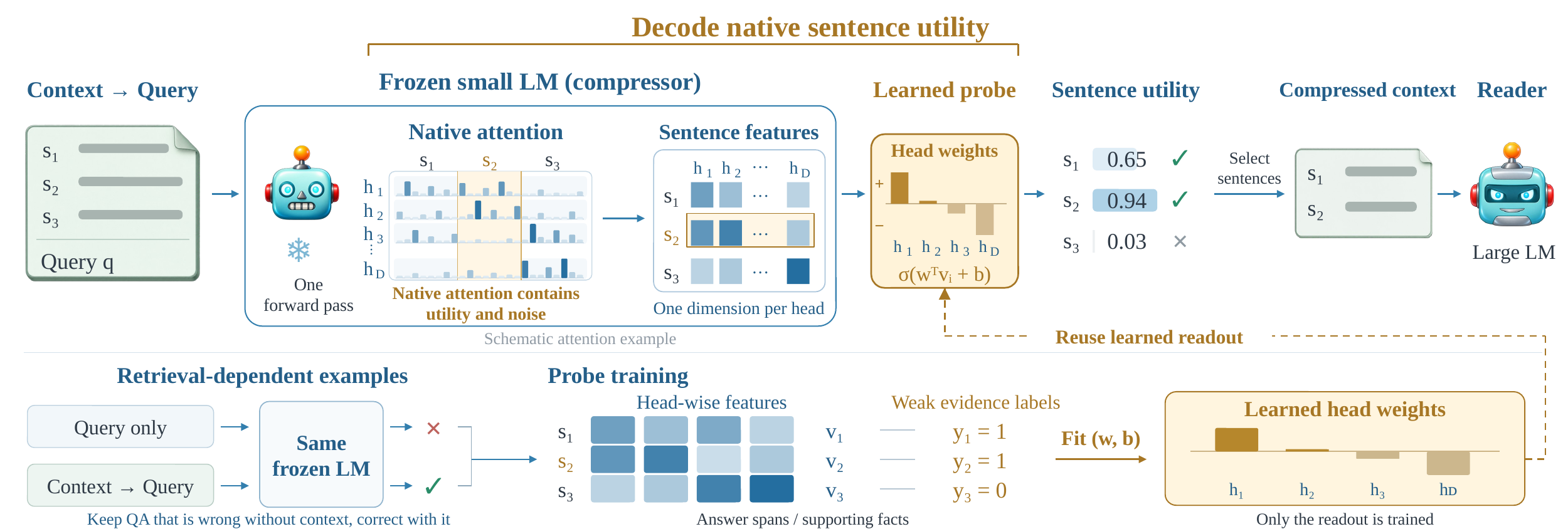}
    \vspace{-3mm} 

\caption{
\textbf{Sentinel Framework Overview.}
Sentinel learns a signed sentence-level readout over a frozen language
model's attention features.
Retrieval-dependent QA selection provides a sample-level behavioral
basis for training; existing answer spans and supporting facts supply
positive sentence labels, with other sentences labeled negative.
Only the probe is trained.
At inference time, one non-autoregressive forward pass scores the
sentences in a native context window for budgeted selection.}
    \label{fig:method-overview}

\end{figure*}

Large language models (LLMs) support a wide range of question-answering
and reasoning tasks \citep{brown2020language,openai2024gpt4technicalreport}.
Retrieval-augmented generation (RAG) extends these capabilities with
evidence from external corpora
\citep{lewis2020retrieval,guu_realm_2020,shi_replug_2023}.
Retrieved contexts, however, can be long, noisy, and redundant,
increasing inference cost and obscuring useful evidence
\citep{liu-etal-2024-lost,yoran2024making_rag_robust}.
This motivates a context compressor that selects query-relevant
information for the downstream model.

Existing methods approach this decision through different importance
signals and training objectives.
Metric-based compressors use predefined importance scores based on
token probabilities or query--context similarity
\citep{jiang_llmlingua_2023,jiang-etal-2024-longllmlingua,li-etal-2023-compressing}.
Several data-driven compressors update their backbone parameters using
labeled examples or generator feedback to learn token- or
sentence-level selection
\citep{pan_llmlingua-2_2024,xu_recomp_2023,hwang_exit_2024}.
Attention-based methods also exploit a language model's internal
representations, including selected retrieval heads and evaluator
heads \citep{wu2024retrieval,fei2025ehpc}.
We investigate a complementary approach: learning how to combine
head-wise signals across layers into a query-aware sentence ranking
while keeping the language model frozen.

A compact language model's contextual representations can support
compression independently of its ability to generate the final answer.
Mechanistic studies associate attention heads with contextual
retrieval and show that their behavior can vary with the input
\citep{wu2024retrieval,jin2024cutting,huang_dynamic_2025,zheng2024attentionheadsurvey}.
These findings motivate treating the head-wise attention profile as a
proxy for query-conditioned context utilization, from which a
lightweight readout can learn sentence utility for compression.
The question is how to read out sentence utility from this
representation with limited adaptation.

We propose \textbf{Sentinel}, a query-aware context compressor that
turns a frozen language model's native attention representations
into sentence-level compression decisions.
A lightweight, signed readout combines each sentence's attention
features across layers and heads to rank it under a token budget.
To ground probe training in the model's answering behavior,
we select retrieval-dependent examples by contrasting
answers produced with and without context.
Existing answer spans and supporting facts supply sentence-level
labels within these examples.
Only the readout is trained; the language model remains frozen.
At inference time, Sentinel scores an entire native context window
in a single non-autoregressive forward pass, without generating an answer.

Experiments on twelve LongBench tasks show that Sentinel can exploit
native attention representations for effective compression with a
0.5B backbone, surpassing Full Context on multiple tasks under a
2K-token budget.
The readout trained on English QA extends to Chinese tasks and
summarization without task-specific retraining, while compressed
contexts remain useful to three downstream readers.
Results across Instruct backbones and the query and signed-weight
ablations further connect this effectiveness to a learned,
query-aware interpretation of head-wise attention signals.

Our contributions are as follows:
\begin{itemize}
\item \textbf{A learned readout of native attention.}
We introduce Sentinel, a query-aware sentence-level compressor that
learns to interpret native attention features across layers and heads
rather than relying on fixed head selection or aggregation.
A signed linear readout maps these features to sentence utility,
using existing English QA annotations while keeping the backbone frozen.

\item \textbf{Effective, transferable, and efficient compression.}
Evaluations on twelve LongBench tasks across three downstream readers
demonstrate effective compression with a 0.5B backbone, surpassing
Full Context on multiple tasks under a 2K-token budget with the Qwen reader.
The same readout transfers to Chinese tasks without task-specific retraining.
Same-hardware profiling further shows lower compression latency and
peak allocated GPU memory than LLMLingua-2 and CPC.

\item \textbf{Empirical evidence for learned sentence selection.}
Query and negative-weight ablations support the contributions of
query conditioning and signed feature weighting.
Sentinel maintains high macro scores across retained fractions,
with a widening advantage over Raw Attention as budgets tighten,
supporting the value of learning to prioritize useful sentences.
\end{itemize}

\section{Methodology}

Sentinel learns a signed linear readout over a frozen language model's
head-wise attention features to estimate query-conditioned sentence utility.
The resulting scores guide sentence selection under a token budget.

\subsection{Context Compression as Contextual Utilization Decoding}
Sentinel reads out query-aware sentence utility from the native
attention patterns of a frozen language model.
Given a query $q$ and a context $C=\{s_1,\ldots,s_n\}$,
it ranks sentences to select a subset $C'\subseteq C$ under a
token budget for a downstream reader.
The query can be a question or a task instruction.
We use contextual utilization decoding to describe the learned mapping
from attention-based proxy features to sentence utility for budgeted
compression.

The compressor jointly processes the query and context using the
prompt in Appendix~\ref{sec:appendix-probingdata}.
Because the final prompt position follows the context, causal
self-attention at that position can attend to all preceding context
tokens \citep{jiang-etal-2024-scaling-prompteol}.
We use its head-wise attention profile as the input to a trained
readout, rather than generating an answer to score the context.

\subsection{Decoding Context Utilization via Probing}
The frozen compressor supplies sentence-level attention features;
a lightweight probe learns their mapping to utility scores.
No language-model parameters are updated.

\subsubsection{Sentence-Level Attention Features}
For each query--context input, we extract attention weights from the
final prompt position across layers and heads.
Let $L$ denote the number of layers and $H$ the number of heads per layer.
The extracted attention rows form
$\mathbf{A} \in \mathbb{R}^{L \times H \times T}$,
where $T$ is the input length and $\mathbf{A}_{l,h,t}$ is the attention
weight from the final prompt position to token $t$ at layer $l$
and head $h$.

For a sentence $s_i$ containing token index set $\mathcal{T}_i$, we compute its normalized attention feature at layer $l$ and head $h$ as
\begin{equation}
a_i^{(l,h)}
=
\frac{
\sum_{t \in \mathcal{T}_i}
\mathbf{A}_{l,h,t}
}{
\sum_{t \in \mathcal{T}_C}
\mathbf{A}_{l,h,t}
}
\label{eq:attention-feature}
\end{equation}  
where $\mathcal{T}_C$ contains the context tokens in the current
input window, excluding prompt, query, and special tokens.
Each feature is the fraction of a head's context-directed attention
assigned to the sentence.

The sentence-level feature vector for sentence $s_i$ is then constructed as
\begin{equation}
\mathbf{v}_i
=
[a_i^{(1,1)}, \dots, a_i^{(L,H)}]
\in
\mathbb{R}^{LH}
\label{eq:sentence-feature}
\end{equation}

\subsubsection{Probing Context Utilization}

We adopt logistic regression as a linear probe that maps each
sentence's head-wise attention profile $\mathbf{v}_i$ to a scalar
utility score for budgeted selection:

\begin{equation}
\hat{y}_i = \sigma(\mathbf{w}^\top \mathbf{v}_i + b)
\end{equation}

where $\mathbf{w}\in\mathbb{R}^{LH}$ is the learned weight vector,
$b$ is the bias, and $\sigma$ is the sigmoid function.
We use $\hat{y}_i\in(0,1)$ to rank sentences.
The weights can be positive or negative, allowing the readout to
combine signals across layers and heads with signed contributions.
The learned weights are fixed at inference time; sentence scores
vary with the query--context attention features.
A linear readout keeps adaptation lightweight and makes its
head-wise weighting explicit.

\subsection{Weak Supervision for Probing Context Utilization}

Probe training combines sample-level behavioral selection with
sentence-level evidence labels from existing QA datasets.
Example selection anchors training in observed context dependence;
answer spans and supporting facts provide sentence supervision.
This requires no additional manual annotation or backbone fine-tuning.

\subsubsection{Retrieval-Dependent Example Selection}

We retain QA instances that meet the
answer-quality criterion with context but fail the closed-book
criterion.
Contrasting these two answering conditions provides a behavioral
basis for selecting training examples.
The dataset-specific EM and F1 thresholds are given in
Appendix~\ref{sec:appendix-probingdata}.

\subsubsection{Sentence-Level Evidence Labeling}

Within the selected examples, sentences containing the gold answer
span receive positive labels.
For HotpotQA, annotated supporting-fact sentences also receive
positive labels, including evidence for intermediate reasoning steps.
Other sentences receive negative labels.
The probe is thus supervised to identify annotated evidence in
contexts selected through the model's answering behavior.

\subsubsection{Sentence Shuffling}

We shuffle sentence order within each training passage to discourage
positional shortcuts \citep{liu-etal-2024-lost}.
Sentence labels are retained under the permutation.

\subsection{Inference-Time Context Compression}

At inference time, Sentinel obtains query-conditioned sentence features
and applies the learned readout in a single non-autoregressive forward
pass per context window.
It greedily packs sentences in descending score order under the
reader-token budget and restores their source order.
The resulting compressed context $C'$ and the original query are
passed to the downstream reader.

\section{Experiments}

\begin{table*}[!t]
\centering
\fontsize{8}{9.4}\selectfont
\setlength{\tabcolsep}{1.8pt}
\renewcommand{\arraystretch}{1.03}
\resizebox{\textwidth}{!}{%
\begin{tabular}{@{}lrrrrrrrrr@{\hspace{4pt}{\color{black!35}\vrule width 0.35pt}\hspace{4pt}}rrrrrrrrr@{}}
\toprule
\multirow{3}{*}{Method}
& \multicolumn{9}{c}{Qwen2.5-7B-Instruct}
& \multicolumn{9}{c}{DeepSeek V4.1 Flash} \\
\cmidrule(lr){2-10}\cmidrule(lr){11-19}
& \multicolumn{4}{c}{English} & \multicolumn{4}{c}{Chinese}
& \multirow{2}{*}{Macro}
& \multicolumn{4}{c}{English} & \multicolumn{4}{c}{Chinese}
& \multirow{2}{*}{Macro} \\
\cmidrule(lr){2-5}\cmidrule(lr){6-9}
\cmidrule(lr){11-14}\cmidrule(lr){15-18}
& SD & MD & Sum & Avg. & SD & MD & Sum & Avg. &
& SD & MD & Sum & Avg. & SD & MD & Sum & Avg. & \\
\midrule
\multicolumn{19}{@{}l}{\textit{Reference conditions}} \\
Empty
& 9.65 & 19.53 & 9.74 & 12.98 & 8.76 & 12.61 & 7.60 & 9.66 & 12.15 & 12.01 & 31.50 & 9.93 & 17.81 & 22.00 & 15.86 & 7.56 & 15.14 & 17.15 \\
\midrule[0.25pt]
\rowcolor{gray!14}
\textbf{Full Context}
& 41.98 & 45.97 & 26.84 & 38.26 & 61.76 & 17.51 & 11.58 & 30.28 & 36.27 & 48.35 & 64.82 & 27.86 & 47.01 & 63.97 & 17.19 & 12.94 & 31.37 & 43.10 \\
\midrule[0.3pt]
\multicolumn{19}{@{}l}{\textit{Metric-Based Context Compression}} \\
LLMLingua-1
& 29.66 & 29.57 & 24.60 & 27.94 & 50.23 & 17.52 & 13.14 & 26.96 & 27.70 & 35.55 & 47.47 & 25.58 & 36.20 & 53.36 & 17.13 & 12.17 & 27.55 & 34.04 \\
LongLLMLingua
& 28.79 & 29.48 & 24.67 & 27.65 & 49.95 & \textbf{18.31} & 11.68 & 26.65 & 27.40 & 35.31 & 49.28 & 25.53 & 36.70 & 52.85 & 16.93 & 12.30 & 27.36 & 34.37 \\
\midrule[0.3pt]
\multicolumn{19}{@{}l}{\textit{Data-Driven Context Compression}} \\
LLMLingua-2
& 29.34 & 30.49 & 25.08 & 28.30 & 49.32 & \underline{17.87} & 12.99 & 26.73 & 27.91 & 34.88 & 50.16 & 25.93 & 36.99 & 53.82 & 17.04 & 12.19 & 27.68 & 34.67 \\
CPC
& 38.41 & 41.45 & 25.28 & 35.05 & 53.64 & 15.97 & 13.48 & 27.70 & 33.21 & 44.22 & 55.55 & 26.71 & 42.16 & 56.26 & 15.88 & 12.59 & 28.24 & 38.68 \\
\midrule[0.3pt]
\multicolumn{19}{@{}l}{\textit{Attention-based compression}} \\
Raw Attention
& 32.41 & 29.65 & 24.48 & 28.85 & 54.62 & 17.54 & 12.86 & 28.34 & 28.72 & 38.84 & 39.95 & 25.52 & 34.77 & 54.57 & \textbf{17.40} & 12.32 & 28.10 & 33.10 \\
Retrieval Head
& 39.33 & 44.50 & 25.82 & 36.55 & \underline{62.31} & 16.94 & 12.26 & 30.50 & 35.04 & 40.78 & 44.42 & 26.41 & 37.20 & 64.16 & \underline{17.39} & 12.11 & 31.22 & 35.71 \\
EHPC
& \underline{40.66} & \textbf{50.99} & 26.39 & \textbf{39.35} & 61.83 & 16.49 & 13.24 & 30.52 & \underline{37.14} & \textbf{48.22} & \textbf{66.34} & \underline{27.71} & \textbf{47.42} & \textbf{65.50} & 16.18 & \textbf{12.92} & \textbf{31.53} & \textbf{43.45} \\
\midrule[0.5pt]
\multicolumn{19}{@{}l}{\textit{Sentinel}} \\
\rowcolor{gray!8}
Sentinel (0.5B)
& 39.66 & 46.12 & \underline{26.40} & \underline{37.39} & \textbf{62.64} & 17.45 & \underline{13.82} & \textbf{31.30} & 35.87 & 43.67 & 60.42 & 27.65 & 43.91 & 64.52 & 16.91 & \underline{12.81} & \underline{31.41} & 40.79 \\
\rowcolor{gray!8}
Sentinel (8B)
& \textbf{41.73} & \underline{49.40} & \textbf{26.91} & \textbf{39.35} & 61.58 & 17.33 & \textbf{14.19} & \underline{31.03} & \textbf{37.27} & \underline{46.74} & \underline{65.66} & \textbf{28.40} & \underline{46.93} & \underline{64.83} & 17.17 & 12.24 & \underline{31.41} & \underline{43.05} \\
\bottomrule
\end{tabular}%
}
\caption{LongBench results by task category and reader.
SD, MD, and Sum denote Single-Document QA, Multi-Document QA,
and Summarization. Avg. averages the nine English or three Chinese
tasks equally; Macro averages all 12 tasks.
Full Context is uncompressed.
DeepSeek thinking is disabled. Best and second-best compression results
in each column are bold and underlined, respectively.}
\label{tab:longbench_readers}
\end{table*}

\subsection{Experimental Setup}

\paragraph{Tasks and metrics.}
We evaluate on 12 LongBench tasks spanning three categories
\citep{bai2024longbench}:
\begin{itemize}
\item \textbf{Single-document QA:} NarrativeQA, Qasper,
MultiFieldQA-en, and MultiFieldQA-zh.
\item \textbf{Multi-document QA:} HotpotQA, 2WikiMQA,
MuSiQue, and DuReader.
\item \textbf{Summarization:} GovReport, QMSum, MultiNews,
and VCSUM.
\end{itemize}
MultiFieldQA-zh, DuReader, and VCSUM are Chinese; the remaining nine
tasks are English.
We evaluate 200 examples per task, except MultiFieldQA-en with 150.

We use the LongBench task prompts and task-specific metrics.
The English QA tasks use answer F1, MultiFieldQA-zh uses Chinese F1,
DuReader and VCSUM use Chinese ROUGE, and the three English
summarization tasks use ROUGE.

\paragraph{Compressors and readers.}
Sentinel uses Qwen2.5-0.5B-Instruct by default and
Llama-3.1-8B-Instruct for the larger-backbone variant.
The readers are Qwen2.5-7B-Instruct, Llama-3.1-8B-Instruct,
and DeepSeek V4.1 Flash.
DeepSeek thinking is disabled.
The model-family study evaluates Instruct checkpoints from Qwen2.5,
Qwen3, Llama-3.2, and Llama-3.1.

\paragraph{Compression protocol.}
Sentinel uses each backbone's native context window.
Over-window documents are scored in maximal sentence-aligned segments
under the same query, followed by document-level sentence selection.
The standard compression budget is 2,000 tokens for all methods.
Sentinel restores selected sentences to their source order.
Token counting and prompt templates are specified in
Appendices~\ref{app:baselines} and~\ref{sec:appendix-probingdata}.
Full Context uses uncompressed text.

\paragraph{Probe training.}
The linear probe is trained on English retrieval-dependent QA examples
from NewsQA, SQuAD, and HotpotQA.
Appendix~\ref{sec:appendix-probingdata} specifies the example selection,
feature extraction, and logistic-regression training procedure.

\paragraph{Baselines.}
We compare three groups of compression baselines:
\begin{itemize}
\item \textbf{Metric-Based.}
LLMLingua-1 \citep{jiang_llmlingua_2023}
and LongLLMLingua \citep{jiang-etal-2024-longllmlingua},
both using Llama-2-7B.
\item \textbf{Data-Driven.}
LLMLingua-2 \citep{pan_llmlingua-2_2024},
using XLM-RoBERTa-large fine-tuned on MeetingBank,
and CPC \citep{liskavets2024promptcpc}, using Mistral-7B-Instruct with LoRA.
\item \textbf{Attention-based.}
Raw Attention and
Retrieval Head \citep{wu2024retrieval},
both using Qwen2.5-0.5B-Instruct;
and EHPC \citep{fei2025ehpc}, using Llama-3.1-8B-Instruct.
Retrieval Head aggregates attention over the top 14 retrieval heads.
\end{itemize}
Empty Context and Full Context serve as reference conditions.
Window settings and implementation details are provided in
Appendix~\ref{app:baselines}.

\subsection{Main Results}

\paragraph{Native attention supports lightweight compression.}
Sentinel's 0.5B results show that useful sentence selection can be
learned from a compact model's frozen attention representations.
Under a 2K-token budget, it exceeds the metric-based and data-driven
compression baselines, Raw Attention, and Retrieval Head on the
English-task average with both Qwen and DeepSeek readers
(Table~\ref{tab:longbench_readers}).
With Qwen, its English average is only 0.87 points below Full Context
(37.39 vs.\ 38.26), while its multi-document QA score slightly
exceeds Full Context (46.12 vs.\ 45.97).
Competitive selection is thus obtained through a lightweight
readout rather than fine-tuning the compression backbone.

\paragraph{Competitive sentence-level compression at 8B.}
Sentinel's 8B variant improves macro scores with both readers and is
competitive with EHPC using the same Llama-3.1-8B-Instruct backbone.
With Qwen, Sentinel achieves the highest compression macro score
of 37.27, exceeding EHPC (37.14) and Full Context (36.27).
With DeepSeek, it reaches 43.05, close to EHPC's 43.45 and
Full Context's 43.10.
Whereas EHPC prunes individual tokens, Sentinel retains whole
sentences, providing explicit selection units that can be inspected
in the original context.

\paragraph{Transfer of the learned selection rule.}
The readout remains useful when both the language and task format
differ from its English QA supervision.
Without additional probe training, Sentinel (0.5B) surpasses Full
Context on the Chinese-task average with Qwen (31.30 vs.\ 30.28).
This transfer also extends to Chinese summarization: on VCSUM,
it scores 13.82 compared with 11.58 for Full Context, despite being
trained only on English QA.

\subsection{Understanding the Learned Readout}

\subsubsection{Effect of the Query}
\begin{table}[t]
\centering
\small
\setlength{\tabcolsep}{7pt}
\begin{tabular}{@{}lr@{}}
\toprule
Compressor query & Answer F1 (\%) \\
\midrule
Correct & \textbf{29.62} \\
Swapped & 6.92 \\
Generic & 3.16 \\
\bottomrule
\end{tabular}
\caption{Query ablation on 100 common valid Qasper QA examples
with a 256-token compression budget and the Qwen2.5-7B-Instruct reader.
The reader always answers the target question.}
\label{tab:query_ablation}
\end{table}
We test whether the learned readout responds to the target information
need rather than assigning query-independent importance to sentences.
On 100 Qasper QA examples, we hold the document and reader question
fixed and use a tight 256-token budget.
Correct supplies the target question.
Swapped supplies another question from the same document, testing
whether document-level topical relevance alone is sufficient.
Generic supplies a summarization instruction, testing whether broadly
informative content can substitute for question-specific evidence.

Correct improves answer F1 by 22.70 and 26.46 points over Swapped
and Generic, respectively (Table~\ref{tab:query_ablation}).
The Swapped result shows that topical alignment with the document
is insufficient: selection depends on the reader's specific
information need.
A generic summarization instruction likewise yields much less useful
context for the target QA task under the same budget.

\subsubsection{Role of Signed Weights}
\begin{table}[t]
\centering
\small
\setlength{\tabcolsep}{4pt}
\begin{tabular}{@{}lrr@{}}
\toprule
Task & Sentinel
& \makecell[r]{Negative weights\\set to zero} \\
\midrule
NarrativeQA     & \textbf{22.73} & 17.03 \\
Qasper          & \textbf{45.19} & 45.19 \\
MultiFieldQA-en & \textbf{51.07} & 50.14 \\
HotpotQA        & \textbf{56.16} & 54.97 \\
2WikiMQA        & \textbf{49.07} & 47.59 \\
MuSiQue         & \textbf{33.12} & 33.06 \\
GovReport       & \textbf{32.32} & 31.76 \\
QMSum           & \textbf{22.72} & 21.65 \\
MultiNews       & \textbf{24.17} & 23.86 \\
MultiFieldQA-zh & \textbf{62.64} & 61.62 \\
DuReader        & \textbf{17.45} & 16.81 \\
VCSUM           & \textbf{13.82} & 12.80 \\
\midrule
Macro           & \textbf{35.87} & 34.71 \\
\bottomrule
\end{tabular}
\caption{Effect of signed probe weights.
Negative coefficients are set to zero without retraining.
Both variants use the Qwen2.5-0.5B-Instruct compressor,
Qwen2.5-7B-Instruct reader, and a 2K-token budget.}
\label{tab:signed_weights}
\end{table}
To examine the contribution of negative weights, we set all negative
probe coefficients to zero without retraining, leaving the positive
coefficients and intercept unchanged.
Both variants use the same attention features, compression budget,
and sentence-packing procedure.
Removing negative coefficients lowers the macro score from 35.87
to 34.71, a decrease of 1.16 points
(Table~\ref{tab:signed_weights}).

The reduction affects eleven tasks, with a tie on Qasper.
The largest drop occurs on NarrativeQA (5.70 points), while declines
also appear in multi-document QA, Chinese QA, and summarization.
For example, QMSum and VCSUM lose 1.07 and 1.02 points, respectively.
The benefit of signed weighting is therefore not confined to a single
task format or language.
Because the attention representation is unchanged, these differences
reflect how the readout combines features into sentence rankings.
Retaining negative coefficients improves downstream performance over
the positive-only readout, supporting joint interpretation of head-wise
signals with signed weights.

\subsubsection{Head-Wise Weight Patterns}
Figure~\ref{fig:head_diagnostics} compares retrieval-head scores
with Sentinel's learned coefficients on the same compressor.
Five of Sentinel's top-14 positively weighted heads overlap with
the top-14 retrieval heads identified by the procedure of
\citet{wu2024retrieval}.
The overlap connects the learned ranking to retrieval-oriented
attention, while the distinct weight pattern shows that Sentinel
does not simply reproduce the selected retrieval-head set.

The probe combines positive and negative contributions across
the full head-wise feature vector.
In a diagnostic sample of 100 HotpotQA examples, several strongly
negative-weighted heads attend predominantly to sink or question
tokens and little to annotated supporting facts
(Appendix Table~\ref{tab:negative_heads}).
These descriptive patterns complement the downstream ablation
in Table~\ref{tab:signed_weights}: attention features outside a
selected positive-head set also participate in the learned readout.

\subsubsection{Attention Feature Configurations}
\begin{table}[t]
\centering
\small
\setlength{\tabcolsep}{5pt}
\begin{tabular}{@{}lr@{}}
\toprule
Configuration & Macro \\
\midrule
\multicolumn{2}{@{}l}{\textit{Attention features}} \\
All-layer (336-d) & \textbf{35.87} \\
Last-layer (14-d) & 34.47 \\
Selected Heads (14-d) & 34.82 \\
\midrule
\multicolumn{2}{@{}l}{\textit{All-layer training configurations}} \\
500 & 35.17 \\
1000 & 34.62 \\
2000 & 34.86 \\
3000 & \textbf{35.87} \\
\bottomrule
\end{tabular}
\caption{Feature and training configurations with the 0.5B compressor,
Qwen reader, and a 2K-token budget.
Task-level scores are given in Appendix~\ref{appendix:ablation-details}.}
\label{tab:readout_configurations}
\end{table}
We compare the full 336-dimensional representation from all 24 layers
with two 14-dimensional subsets: Last-layer uses the final-layer heads,
while Selected Heads uses heads selected across layers with mRMR
\citep{ding2005minimum}.
Each configuration trains a sentence-level readout and uses the same
0.5B backbone, Qwen reader, and 2K-token output budget.
The all-layer representation achieves the best macro score, exceeding
the Last-layer and Selected Heads variants by 1.40 and 1.05 points, respectively
(Table~\ref{tab:readout_configurations}).
Retaining the full head-wise profile therefore offers the strongest
compression among the evaluated configurations.
At the same 14-dimensional feature budget, Selected Heads scores 0.35 points
above Last-layer.
Nevertheless, the restricted inputs preserve much of this performance
with only one twenty-fourth as many features,
showing that a compact attention representation already contains
substantial information for sentence selection.

\subsection{Generality and Supervision Requirements}
\subsubsection{Compressor Families and Scales}
\label{sec:proxy-scale-analysis}
\begin{figure}[t]
\centering
\includegraphics[width=\linewidth]{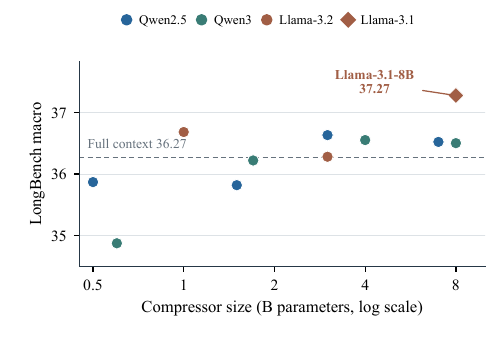}
\caption{LongBench macro scores across Instruct compressor families
with the Qwen reader and a 2K-token budget.
The highest-scoring checkpoint is labeled; the dashed line marks Full Context.}
\label{fig:proxy_family_size}
\end{figure}
We train a separate readout for each of eleven Instruct checkpoints
from Qwen2.5, Qwen3, Llama-3.2, and Llama-3.1, spanning 0.5B--8B
parameters (Figure~\ref{fig:proxy_family_size}).
All checkpoints yield useful compression, with macro scores ranging
from 34.88 to 37.27 (Table~\ref{tab:longbench_models}).
The 3B--8B Qwen backbones improve on the smallest checkpoint in each
family, and all four exceed the Full Context
macro score of 36.27.
The highest score, 37.27, is obtained by Llama-3.1-8B-Instruct,
exceeding Full Context by 1.00 point.
Together with the strong Llama-3.2-1B result, these findings show
that the readout applies across model families, with larger backbones
offering further gains while compact models remain effective.

\subsubsection{Probing Data Configurations}
We evaluate the 500, 1000, 2000, and 3000 probe-training configurations
using the same 336-dimensional all-layer features, 0.5B backbone,
Qwen reader, and 2K-token budget.
Only the linear readout is trained; the backbone remains frozen.
The 500 configuration already reaches 35.17 macro points, within
0.70 points of the best 3000 configuration
(Table~\ref{tab:readout_configurations}).
Although the intermediate configurations vary, all four retain
effective compression, indicating that limited supervision suffices
to make the pretrained model's attention features useful for selection.

\subsection{Compression Trade-offs}
\subsubsection{Compression-Ratio Variants}
\begin{figure}[t]
\centering
\includegraphics[width=\linewidth]{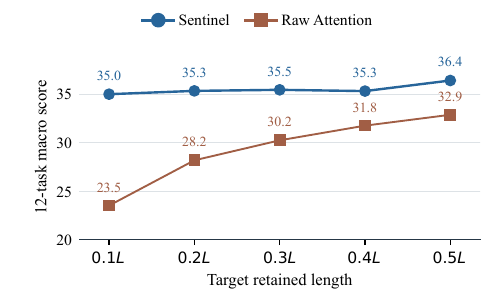}
\caption{Twelve-task macro scores by target retained length.
$L$ is each example's original context length; the ticks denote
target budgets of $0.1L$--$0.5L$.}
\label{fig:compression_ratio}
\end{figure}
Figure~\ref{fig:compression_ratio} tests sentence-ranking quality as the
target retained fraction decreases, using the same 12 tasks at every point.
Because sentences are packed in descending score order, tight budgets
test whether useful information is concentrated among the
highest-scoring sentences.

Sentinel maintains high macro scores with little variation across the
tested fractions, whereas Raw Attention's performance falls as less
context is retained.
At the most aggressive target, $0.1L$, Sentinel scores 35.00 and
exceeds Raw Attention by 11.47 points.
This contrast indicates that the trained readout learns to place
task-useful sentences near the top of its ranking.
For Raw Attention, larger budgets can compensate for weaker
selectivity by retaining useful sentences assigned lower scores,
masking ranking errors that become costly under tight budgets.

\subsubsection{Compression Cost}
\begin{figure*}[t]
\centering
\includegraphics[width=\textwidth]{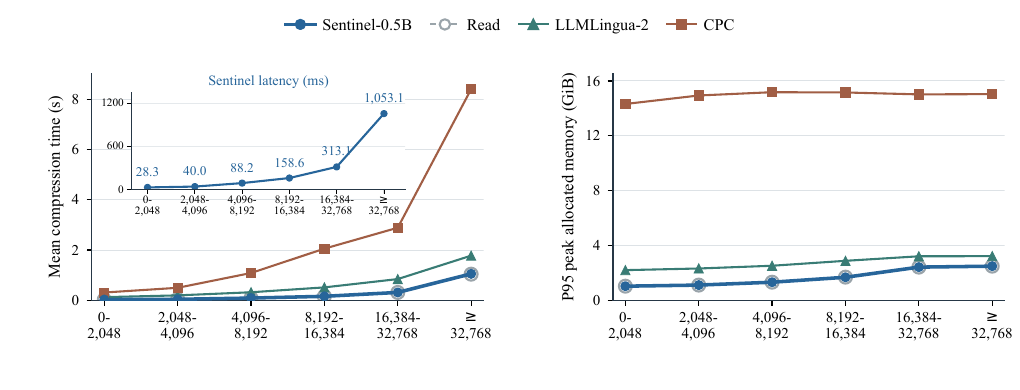}
\caption{Compressor-only cost by input-document length. Left: mean
\texttt{compress()} latency in seconds; the inset shows Sentinel's
latency in milliseconds. Right: P95 peak allocated GPU memory.
All methods were profiled on the same NVIDIA RTX 5090 GPU.
Bins reflect pre-compression document length,
not necessarily the length attended by each method. Input-length bins
are given in tokens; the edge bins are sparsely sampled.
Appendix~\ref{sec:efficiency} gives the profiling protocol.}
\label{fig:length_compression_cost}
\end{figure*}

We measure compressor-only cost on a fixed set of 120 LongBench examples
(10 per task), stratified by input-document length.
Sentinel, Read, LLMLingua-2, and CPC are profiled under the same
measurement protocol on the same machine equipped with an
NVIDIA RTX 5090 GPU.
Length is counted on the document/context with the
Qwen2.5-7B-Instruct tokenizer.
Figure~\ref{fig:length_compression_cost} compares compressor-only
latency and the 95th percentile of peak allocated GPU memory,
including resident compressor weights.
Reader inference is excluded.

Sentinel has lower recorded latency and peak allocation than
LLMLingua-2 and CPC across the reported length bins.
In the 8,192--16,384-token input-length bin, its mean compression latency
is 158.6 ms, with a P95 peak allocation of 1.68 GiB.
At this length, Sentinel is 3.2 times faster than LLMLingua-2 and
13.0 times faster than CPC, while using substantially less peak memory.
Read is an implementation reference on the same 0.5B probe/SDPA
path and nearly overlaps with Sentinel.
Appendix~\ref{sec:efficiency} gives the profiling protocol.

\section{Related Work}

\paragraph{Metric-Based Context Compression}
Metric-based approaches select tokens or sentences using predefined
importance scores, such as self-information, mutual information,
or query--context similarity.
LLMLingua-1 \citep{jiang_llmlingua_2023} and LongLLMLingua
\citep{jiang-etal-2024-longllmlingua} prune tokens based on
perplexity or query-conditioned probability estimates.
Selective Context \citep{li-etal-2023-compressing} removes
low-information content using token-level self-information.
These approaches apply a specified scoring rule without training
a dedicated compression model.

\paragraph{Data-Driven Context Compression}
Data-driven approaches learn compression decisions from external supervision, typically by training a ranking or classification model to predict which tokens or sentences should be retained. Token-level methods such as LLMLingua-2 \citep{pan_llmlingua-2_2024} leverage distilled labels from large language models to train lightweight compressors. 
At the sentence level, methods such as RECOMP \citep{xu_recomp_2023} train compressors to produce extractive or abstractive summaries that improve downstream performance, while EXIT \citep{hwang_exit_2024} learns a sentence-level classifier to select query-relevant sentences.
Other works, such as CPC \citep{liskavets2024promptcpc}, Refiner \citep{li_refiner_2024}, and FineFilter \citep{zhang_finefilter_2025}, further incorporate query-aware ranking, structure-aware reranking, or multi-hop reasoning objectives.
Sentinel also learns a compression rule, but trains only a linear
readout over a frozen backbone's head-wise attention features,
using existing QA evidence labels within retrieval-dependent examples.

\paragraph{Attention-Based Context Compression}

Many attention-based compressors select context through predefined
attention aggregation, head selection, or combinations with other
importance signals.
QUITO \citep{wang_quito_2024} and AttentionRAG
\citep{fang_attentionrag_2025} aggregate query--context attention
to rank and filter context spans.
GemFilter \citep{shi2024gemfilter} uses attention in early layers to
select input tokens, then processes the reduced sequence with the
full model.
EHPC \citep{fei2025ehpc} performs token-level compression using
selected evaluator heads, observation-window attention averaging,
and local pooling.
DAC \citep{zhao-etal-2025-dac} combines attention with token information
entropy for iterative token pruning.
\citet{zhao2025attncomp} take the maximum
query-to-context attention over selected retrieval heads and group
tokens into semantic units using a self-attention graph.
Beyond directly reusing native attention, \citet{luo-etal-2025-attncomp}
train an additional cross-attention layer and apply adaptive Top-$P$
selection at the document level.

Retrieval heads can be dynamically activated by the input
\citep{wu2024retrieval}, and different heads can play opposing roles
in relying on contextual versus parametric knowledge
\citep{jin2024cutting}.
Such input-dependent and potentially opposing roles motivate learning
a readout of the full head-wise profile, rather than simply aggregating
selected heads.
Sentinel learns a signed linear readout across layers and heads,
allowing head-wise features to increase or suppress sentence scores
while keeping the backbone frozen.
It retains original sentences rather than individual tokens or entire
documents.

\section{Conclusion}

We introduced Sentinel, which learns a signed sentence-level readout
of a frozen language model's native attention representations.
The method jointly interprets head-wise features across layers rather
than relying on direct attention aggregation.
A 0.5B backbone surpasses Full Context on multiple LongBench tasks
under a 2K-token budget, while an 8B backbone improves overall
performance with both main-table readers.
The English-QA-trained readout extends to Chinese tasks and
summarization, and its selections remain useful across three readers.
Query and negative-weight ablations support the importance of
conditioning and signed contributions.
These results support the practical value of native attention features
as a proxy for contextual utilization, enabling effective query-aware
compression through a lightweight learned readout.

\clearpage

\section*{Limitations}
\paragraph{Scope of supervision.}
Sentinel uses existing English QA annotations to train its
lightweight readout, requiring no additional manual labeling.
Despite this restricted supervision, the same readout transfers
effectively to Chinese tasks and summarization.
Our study focuses on establishing the utility of this readily
available supervision; we have not systematically explored alternative
supervision sources or multilingual and task-diverse training mixtures.
Whether such extensions yield further gains remains an open question.

\paragraph{Compression budget selection.}
Our experiments evaluate Sentinel under both a fixed token
budget and different target retained fractions.
The budget is specified externally, providing direct control over the
amount of context passed to the reader.
We have not explored automatic budget selection based on the query
and document characteristics.
Learning to adapt the budget to individual inputs while balancing
answer quality and inference cost remains a direction for future work.

\section*{AI Assistance Statement}

Generative AI tools, including ChatGPT and Gemini, were used to assist with language editing during the preparation of this work. All generated content was reviewed and verified by the authors. The authors take full responsibility for the accuracy, integrity, and originality of the final manuscript.

\bibliography{custom}

\clearpage
\appendix
\setcounter{dbltopnumber}{4}
\renewcommand{\dbltopfraction}{0.90}
\renewcommand{\textfraction}{0.05}
\makeatletter
\setlength{\@dblfptop}{0pt}
\setlength{\@dblfpsep}{12pt plus 2pt minus 2pt}
\makeatother
\raggedbottom
\section{Additional Probing Data and Training Details}
\label{sec:appendix-probingdata}

\paragraph{Probing data.}
The original 0.5B probe samples NewsQA (50\%), SQuAD (20\%),
and HotpotQA (30\%).
Each accepted QA example supplies positive and negative sentence labels.
We segment retrieved contexts with spaCy's \texttt{sentencizer}
and use the same boundaries for supervision and attention aggregation.
Under the Qwen2.5 tokenizer, 30.1\% of the NewsQA training contexts
contain at most 500 tokens and 69.9\% contain 500--1,000;
99.3\% of SQuAD and all HotpotQA training contexts contain at most 500.
Additional feature and probe-training configurations are reported in
Appendix~\ref{appendix:ablation-details}.

We use the original HotpotQA contexts for probe training;
LongBench evaluates an extended-context version of the task.
We exclude training examples whose questions appear in LongBench.

\paragraph{Compressor prompt.}
We extract sentence-level attention features with the following
query--context prompt:

\begin{tcolorbox}[colback=gray!5!white, colframe=gray!50!black, boxrule=0.2pt]
\small
\texttt{Given the following information: \{context\}} \\
\texttt{Answer the following question based on the given information with one or a few words: \{question\}} \\
\texttt{Answer:}
\end{tcolorbox}

The \texttt{\{question\}} field contains the QA question or QMSum query.
For GovReport, MultiNews, and VCSUM it contains
\texttt{Summarize the document.}
Readers use their corresponding LongBench task templates.

\paragraph{Example selection thresholds.}
For NewsQA and SQuAD, the acceptance thresholds are closed-book EM = 0
and context-provided EM = 1.
For HotpotQA, we require closed-book F1 $\leq 0.2$
and context-provided F1 $\geq 0.5$.

\paragraph{Probe optimization.}
We train logistic regression with class-balanced weighting,
$\ell_2$ regularization, and the liblinear solver.
Five-fold cross-validation selects $C$ from
$\{0.01,0.1,1,10,100\}$ by validation AUC;
optimization is capped at 2,000 iterations.

\section{Baseline Implementation and Reader Settings}
\label{app:baselines}

\paragraph{Compressor windows and budgets.}
The native input limits are 32,768 tokens for Qwen2.5
(0.5B, 1.5B, 3B, and 7B), 40,960 for Qwen3
(0.6B, 1.7B, 4B, and 8B), and 131,072 for
Llama-3.2 (1B and 3B) and Llama-3.1 (8B).
Retrieval Head retains the final 32,768 tokens of over-window inputs
before attention extraction.
API-reader results use the reported 2K-token compressed contexts;
Retrieval Head reuses the contexts packed with the Qwen reader's tokenizer.

\paragraph{Sentinel windowing and packing.}
Attention extraction uses only the row at the final prompt position,
rather than the full token-by-token attention matrix.
The implementation uses the final-row extraction path with scaled
dot-product attention (SDPA).
We reserve space for the query, template, and special tokens within
the compressor's native input window.
Longer contexts are divided at sentence boundaries into maximal
segments, each conditioned on the same query or instruction.
The same probe scores every segment; its sigmoid sentence scores
are directly pooled into a document-level candidate list.
The final, shorter segment is processed identically, without
additional adjustment.
We consider sentences in descending score order and accept each
sentence only if the re-tokenized selected context remains within
the reader-token budget; sentences that do not fit are skipped.
Selected sentences are restored to source order.
There is one output budget for the document, not one per segment.

\begin{itemize}
    \item \textbf{LLMLingua-1 and LongLLMLingua} both use
    \texttt{NousResearch/}\allowbreak\texttt{Llama-2-7b-hf}.

    \item \textbf{LLMLingua-2} uses the
    XLM-RoBERTa-large MeetingBank checkpoint;
    \textbf{CPC} uses Mistral-7B-Instruct with CPC LoRA.

    \item \textbf{Raw Attention} directly aggregates attention
    from the same 0.5B compressor and window as Sentinel.

    \item \textbf{Retrieval Head} directly aggregates attention over
    the top 14 retrieval heads of Qwen2.5-0.5B-Instruct.

    \item \textbf{EHPC} \citep{fei2025ehpc} uses the official implementation
    with Llama-3.1-8B-Instruct as the compressor,
    layer 13, and heads $[18,13,21,8,11,1,4,3]$.
    The implementation is distributed with the paper's supplementary material.

\end{itemize}

\paragraph{Reader generation.}
For local readers, we use the LongBench task prompts
\citep{bai2024longbench} with temperature 0, $top_p=1$,
seed 42, and one non-streaming generation per example.

\section{Reader Task-Level Results}
\label{app:reader_tasks}
\label{app:deepseek}
Table~\ref{tab:reader_tasks} reports task-level results for the Qwen
and DeepSeek readers in Table~\ref{tab:longbench_readers},
as well as the additional Llama reader.
\begin{table*}[t]
\centering
\scriptsize
\setlength{\tabcolsep}{3pt}
\resizebox{\linewidth}{!}{
\begin{tabular}{lrrrrrrrrrrrrrrr}
\toprule
Method & NQA & Qasper & MF-en & Hotpot & 2Wiki & MuSiQue
& GovReport & QMSum & MultiNews & MF-zh & DuReader & VCSUM & EN Avg. & ZH Avg. & Macro \\
\midrule
\multicolumn{16}{c}{\textit{Reader: Qwen2.5-7B-Instruct}} \\
\midrule
Empty
& 3.77 & 11.00 & 14.19 & 25.24 & 24.77 & 8.58 & 6.31 & 13.56 & 9.36 & 8.76 & 12.61 & 7.60 & 12.98 & 9.66 & 12.15 \\
\rowcolor{gray!14}
\textbf{Full Context}
& 29.67 & 44.16 & 52.12 & 59.30 & 47.31 & 31.29 & 32.69 & 23.58 & 24.24 & 61.76 & 17.51 & 11.58 & 38.26 & 30.28 & 36.27 \\
LLMLingua-1
& 10.52 & 33.05 & 45.41 & 34.55 & 36.37 & 17.79 & 29.86 & 19.57 & 24.37 & 50.23 & 17.52 & 13.14 & 27.94 & 26.96 & 27.70 \\
LongLLMLingua
& 10.30 & 32.75 & 43.32 & 33.60 & 34.35 & 20.49 & 29.99 & 19.90 & 24.11 & 49.95 & 18.31 & 11.68 & 27.65 & 26.65 & 27.40 \\
LLMLingua-2
& 11.96 & 35.53 & 40.52 & 35.02 & 38.50 & 17.95 & 29.79 & 21.71 & 23.74 & 49.32 & 17.87 & 12.99 & 28.30 & 26.73 & 27.91 \\
CPC
& 21.14 & 44.06 & 50.04 & 51.67 & 41.96 & 30.71 & 29.62 & 22.27 & 23.94 & 53.64 & 15.97 & 13.48 & 35.05 & 27.70 & 33.21 \\
Raw Attention
& 9.71 & 40.63 & 46.89 & 35.46 & 34.84 & 18.66 & 29.84 & 19.72 & 23.88 & 54.62 & 17.54 & 12.86 & 28.85 & 28.34 & 28.72 \\
Retrieval Head
& 19.25 & 46.83 & 51.92 & 54.14 & 47.66 & 31.71 & 31.94 & 21.77 & 23.76 & 62.31 & 16.94 & 12.26 & 36.55 & 30.50 & 35.04 \\
EHPC
& 25.99 & 45.58 & 50.40 & 64.64 & 49.48 & 38.85 & 32.05 & 23.22 & 23.90 & 61.83 & 16.49 & 13.24 & 39.35 & 30.52 & 37.14 \\
\rowcolor{gray!8}
Sentinel (0.5B)
& 22.73 & 45.19 & 51.07 & 56.16 & 49.07 & 33.12 & 32.32 & 22.72 & 24.17 & 62.64 & 17.45 & 13.82 & 37.39 & 31.30 & 35.87 \\
\rowcolor{gray!8}
Sentinel (8B)
& 29.02 & 45.47 & 50.70 & 60.92 & 51.25 & 36.02 & 32.97 & 23.64 & 24.12 & 61.58 & 17.33 & 14.19 & 39.35 & 31.03 & 37.27 \\
\midrule
\multicolumn{16}{c}{\textit{Reader: Llama-3.1-8B-Instruct}} \\
\midrule
Empty
& 7.52 & 10.50 & 6.04 & 25.05 & 17.23 & 8.57 & 15.18 & 12.90 & 9.01 & 19.51 & 11.52 & 7.37 & 12.44 & 12.80 & 12.53 \\
\rowcolor{gray!14}
\textbf{Full Context}
& 28.04 & 45.22 & 55.99 & 57.42 & 47.91 & 33.24 & 34.53 & 25.19 & 27.09 & 63.15 & 24.58 & 11.64 & 39.40 & 33.12 & 37.83 \\
LLMLingua-1
& 14.91 & 30.95 & 45.72 & 39.58 & 33.40 & 21.35 & 31.35 & 20.46 & 26.72 & 51.04 & 20.37 & 13.83 & 29.38 & 28.41 & 29.14 \\
LongLLMLingua
& 14.16 & 32.23 & 46.09 & 40.34 & 34.67 & 22.37 & 30.78 & 20.89 & 26.61 & 52.52 & 20.91 & 13.75 & 29.79 & 29.06 & 29.61 \\
LLMLingua-2
& 17.86 & 34.71 & 40.10 & 41.84 & 39.89 & 19.20 & 30.07 & 22.42 & 26.56 & 52.12 & 20.36 & 13.76 & 30.29 & 28.75 & 29.90 \\
CPC
& 25.49 & 41.61 & 53.70 & 49.26 & 44.17 & 30.61 & 29.47 & 24.00 & 26.02 & 57.74 & 19.75 & 14.48 & 36.04 & 30.66 & 34.69 \\
Raw Attention
& 19.49 & 44.85 & 53.83 & 52.38 & 42.48 & 28.58 & 31.85 & 22.05 & 26.16 & 59.54 & 22.37 & 13.36 & 35.74 & 31.76 & 34.75 \\
Retrieval Head
& 20.31 & 45.27 & 53.56 & 56.45 & 45.90 & 31.93 & 32.76 & 23.50 & 26.48 & 61.60 & 20.38 & 13.30 & 37.35 & 31.76 & 35.95 \\
EHPC
& 28.06 & 44.91 & 51.02 & 59.31 & 53.20 & 39.60 & 30.82 & 24.53 & 26.25 & 63.72 & 19.07 & 12.60 & 39.74 & 31.80 & 37.76 \\
\rowcolor{gray!8}
Sentinel (0.5B)
& 20.69 & 46.51 & 54.68 & 57.60 & 41.28 & 30.38 & 32.67 & 24.07 & 26.48 & 64.24 & 20.37 & 14.09 & 37.15 & 32.90 & 36.09 \\
\rowcolor{gray!8}
Sentinel (8B)
& 28.16 & 46.14 & 54.23 & 58.29 & 48.46 & 36.63 & 33.14 & 25.08 & 26.50 & 63.47 & 20.20 & 13.72 & 39.63 & 32.46 & 37.83 \\
\midrule
\multicolumn{16}{c}{\textit{Reader: DeepSeek V4.1 Flash (thinking off)}} \\
\midrule
Empty
& 8.75 & 11.00 & 16.27 & 38.51 & 37.98 & 18.01 & 4.32 & 17.77 & 7.71 & 22.00 & 15.86 & 7.56 & 17.81 & 15.14 & 17.15 \\
\rowcolor{gray!14}
\textbf{Full Context}
& 38.68 & 47.55 & 58.82 & 70.45 & 70.78 & 53.24 & 32.72 & 23.80 & 27.05 & 63.97 & 17.19 & 12.94 & 47.01 & 31.37 & 43.10 \\
LLMLingua-1
& 25.41 & 32.85 & 48.40 & 54.35 & 55.52 & 32.53 & 30.54 & 19.38 & 26.82 & 53.36 & 17.13 & 12.17 & 36.20 & 27.55 & 34.04 \\
LongLLMLingua
& 24.64 & 33.52 & 47.77 & 56.60 & 56.45 & 34.78 & 30.50 & 19.34 & 26.74 & 52.85 & 16.93 & 12.30 & 36.70 & 27.36 & 34.37 \\
LLMLingua-2
& 24.43 & 37.62 & 42.58 & 56.70 & 59.83 & 33.94 & 30.63 & 20.63 & 26.54 & 53.82 & 17.04 & 12.19 & 36.99 & 27.68 & 34.67 \\
CPC
& 30.57 & 43.95 & 58.14 & 62.03 & 63.44 & 41.19 & 30.89 & 22.51 & 26.74 & 56.26 & 15.88 & 12.59 & 42.16 & 28.24 & 38.68 \\
Raw Attention
& 23.29 & 42.58 & 50.66 & 44.68 & 53.51 & 21.67 & 30.62 & 19.16 & 26.79 & 54.57 & 17.40 & 12.32 & 34.77 & 28.10 & 33.10 \\
Retrieval Head
& 24.67 & 45.70 & 51.97 & 47.75 & 65.12 & 20.38 & 32.34 & 19.81 & 27.08 & 64.16 & 17.39 & 12.11 & 37.20 & 31.22 & 35.71 \\
EHPC
& 35.35 & 50.80 & 58.50 & 69.98 & 74.57 & 54.46 & 33.09 & 23.31 & 26.72 & 65.50 & 16.18 & 12.92 & 47.42 & 31.53 & 43.45 \\
\rowcolor{gray!8}
Sentinel (0.5B)
& 29.47 & 48.27 & 53.26 & 67.08 & 65.61 & 48.58 & 34.16 & 21.75 & 27.05 & 64.52 & 16.91 & 12.81 & 43.91 & 31.41 & 40.79 \\
\rowcolor{gray!8}
Sentinel (8B)
& 34.85 & 47.32 & 58.04 & 71.11 & 71.50 & 54.38 & 34.62 & 23.79 & 26.79 & 64.83 & 17.17 & 12.24 & 46.93 & 31.41 & 43.05 \\
\bottomrule
\end{tabular}
}
\caption{Task-level results across three readers.
EN Avg. and ZH Avg. average the nine English and three Chinese tasks,
respectively; Macro averages all 12 tasks equally.}
\label{tab:reader_tasks}
\label{tab:deepseek_tasks}
\end{table*}

\section{Additional Results on Compressor Model Size and Family}
\label{sec:appendix-scale}
All results in Table~\ref{tab:longbench_models} use the Qwen reader.
Each backbone has its own all-layer probe, following Sentinel's
QA data construction and probe-training procedure
(Appendix~\ref{sec:appendix-probingdata}).
Native context windows are listed in Appendix~\ref{app:baselines}.
The Llama-3.1-8B entry uses the Sentinel (8B) configuration and results
from Table~\ref{tab:longbench_readers}.
\begin{table*}[t]
\centering
\scriptsize
\setlength{\tabcolsep}{3pt}
\resizebox{\linewidth}{!}{
\begin{tabular}{lrrrrrrrrrrrrrrr}
\toprule
Method & NQA & Qasper & MF-en & Hotpot & 2Wiki & MuSiQue
& GovReport & QMSum & MultiNews & MF-zh & DuReader & VCSUM & EN Avg. & ZH Avg. & Macro \\
\midrule
Qwen2.5-0.5B
& 22.73 & 45.19 & 51.07 & 56.16 & 49.07 & 33.12 & 32.32 & 22.72 & 24.17 & 62.64 & 17.45 & 13.82 & 37.39 & 31.30 & 35.87 \\

Qwen2.5-1.5B
& 22.17 & 46.80 & 52.07 & 58.52 & 46.89 & 32.97 & 32.85 & 23.42 & 24.31 & 59.96 & 16.55 & 13.31 & 37.78 & 29.94 & 35.82 \\

Qwen2.5-3B
& 24.37 & 47.01 & 51.15 & 61.17 & 49.97 & 36.58 & 32.33 & 23.43 & 24.32 & 59.58 & 16.90 & 12.70 & 38.93 & 29.73 & 36.63 \\

Qwen2.5-7B
& 25.20 & 44.79 & 52.45 & 60.45 & 47.59 & 35.86 & 32.94 & 23.93 & 24.09 & 60.65 & 17.26 & 13.04 & 38.59 & 30.32 & 36.52 \\

Qwen3-0.6B
& 22.52 & 45.47 & 52.16 & 49.40 & 48.82 & 30.33 & 32.23 & 21.87 & 23.76 & 61.04 & 17.38 & 13.53 & 36.28 & 30.65 & 34.88 \\

Qwen3-1.7B
& 23.09 & 44.50 & 51.62 & 58.65 & 52.16 & 33.24 & 32.32 & 23.45 & 24.04 & 61.46 & 17.07 & 13.09 & 38.12 & 30.54 & 36.22 \\

Llama-3.2-1B
& 29.16 & 45.47 & 51.76 & 57.38 & 50.73 & 34.76 & 32.79 & 22.81 & 24.00 & 61.11 & 17.65 & 12.59 & 38.76 & 30.45 & 36.68 \\

Qwen3-4B
& 24.10 & 44.81 & 51.82 & 60.27 & 49.97 & 38.20 & 32.50 & 23.38 & 24.13 & 59.80 & 16.91 & 12.76 & 38.80 & 29.82 & 36.55 \\

Llama-3.2-3B
& 26.79 & 45.36 & 50.36 & 57.77 & 50.73 & 34.08 & 33.00 & 23.14 & 24.36 & 59.79 & 17.25 & 12.71 & 38.40 & 29.92 & 36.28 \\

Llama-3.1-8B
& 29.02 & 45.47 & 50.70 & 60.92 & 51.25 & 36.02 & 32.97 & 23.64 & 24.12 & 61.58 & 17.33 & 14.19 & 39.35 & 31.03 & 37.27 \\

Qwen3-8B
& 24.24 & 46.57 & 51.70 & 58.70 & 50.21 & 35.51 & 32.17 & 23.67 & 24.14 & 61.79 & 17.04 & 12.25 & 38.55 & 30.36 & 36.50 \\
\bottomrule
\end{tabular}
}
\caption{Task-level Sentinel scores across Instruct compressor families with the Qwen reader.
EN Avg. and ZH Avg. average the nine English and three Chinese tasks,
respectively; Macro averages all 12 tasks equally.}
\label{tab:longbench_models}
\end{table*}

\section{Ablation Details}
\label{appendix:ablation-details}
\paragraph{Attention features.}
Last-layer uses the 14 final-layer heads;
Selected Heads uses 14 heads selected from across layers.
Both follow Sentinel's QA data construction and probe-training procedure.
\begin{table*}[t]
\centering
\scriptsize
\setlength{\tabcolsep}{3pt}
\resizebox{\linewidth}{!}{
\begin{tabular}{lrrrrrrrrrrrrrrr}
\toprule
Method & NQA & Qasper & MF-en & Hotpot & 2Wiki & MuSiQue
& GovReport & QMSum & MultiNews & MF-zh & DuReader & VCSUM & EN Avg. & ZH Avg. & Macro \\
\midrule
All-layer (336-d)
& 22.73 & 45.19 & 51.07 & 56.16 & 49.07 & 33.12 & 32.32 & 22.72 & 24.17 & 62.64 & 17.45 & 13.82 & 37.39 & 31.30 & 35.87 \\
\midrule
\multicolumn{16}{l}{\textit{Feature configurations}} \\
Last-layer (14-d)
& 20.28 & 45.21 & 50.69 & 54.10 & 46.07 & 31.16 & 31.96 & 21.46 & 23.85 & 58.60 & 17.34 & 12.88 & 36.09 & 29.61 & 34.47 \\
Selected Heads (14-d)
& 20.50 & 45.01 & 50.80 & 55.33 & 47.41 & 30.54 & 32.13 & 21.71 & 23.76 & 60.55 & 16.65 & 13.50 & 36.35 & 30.23 & 34.82 \\
\midrule
\multicolumn{16}{l}{\textit{Training configurations (all-layer)}} \\
500
& 18.11 & 46.07 & 52.05 & 55.95 & 48.84 & 32.61 & 32.62 & 21.41 & 23.91 & 60.38 & 17.14 & 13.00 & 36.84 & 30.17 & 35.17 \\

1000
& 17.19 & 45.14 & 50.41 & 54.26 & 47.25 & 32.16 & 32.39 & 21.72 & 23.82 & 61.06 & 17.05 & 13.02 & 36.04 & 30.38 & 34.62 \\

2000
& 18.40 & 45.55 & 51.19 & 54.47 & 47.09 & 31.13 & 32.35 & 21.99 & 23.75 & 62.05 & 16.73 & 13.65 & 36.21 & 30.81 & 34.86 \\

\bottomrule
\end{tabular}
}
\caption{Attention feature and probe-training configurations with the Qwen reader.
EN Avg. and ZH Avg. average the nine English and three Chinese tasks,
respectively; Macro averages all 12 tasks equally.}
\label{tab:feature_ablation}
\label{tab:training_sample_sizes}
\end{table*}

\begin{table*}[t]
\centering
\scriptsize
\setlength{\tabcolsep}{3pt}
\resizebox{\linewidth}{!}{%
\begin{tabular}{lrrrrrrrrrr}
\toprule
& \multicolumn{5}{c}{Raw Attention} & \multicolumn{5}{c}{Sentinel} \\
\cmidrule(lr){2-6}\cmidrule(lr){7-11}
Task & $0.1L$ & $0.2L$ & $0.3L$ & $0.4L$ & $0.5L$
     & $0.1L$ & $0.2L$ & $0.3L$ & $0.4L$ & $0.5L$ \\
\midrule
NarrativeQA & 14.03 & 15.79 & 19.68 & 23.47 & 20.81 & 13.79 & 21.92 & 21.44 & 23.24 & 28.01 \\
Qasper & 28.46 & 35.47 & 40.55 & 42.42 & 41.81 & 44.65 & 45.85 & 45.90 & 46.10 & 46.56 \\
MultiFieldQA-en & 37.21 & 45.31 & 47.03 & 47.91 & 48.65 & 48.20 & 48.71 & 49.21 & 49.57 & 51.20 \\
HotpotQA & 31.66 & 36.10 & 41.59 & 41.27 & 46.45 & 58.90 & 58.60 & 59.84 & 57.37 & 58.92 \\
2WikiMQA & 27.64 & 35.35 & 35.75 & 39.22 & 40.35 & 49.23 & 46.77 & 45.05 & 44.22 & 50.82 \\
MuSiQue & 16.02 & 20.98 & 21.85 & 23.60 & 27.49 & 31.91 & 30.30 & 31.50 & 29.81 & 31.31 \\
GovReport & 27.66 & 29.92 & 30.65 & 31.07 & 31.48 & 32.46 & 32.31 & 32.14 & 32.60 & 32.77 \\
QMSum & 18.65 & 19.99 & 21.27 & 21.48 & 22.50 & 23.85 & 23.26 & 23.67 & 23.34 & 23.36 \\
MultiNews & 19.42 & 21.21 & 22.17 & 22.80 & 23.16 & 24.15 & 24.03 & 23.57 & 23.70 & 23.15 \\
MultiFieldQA-zh & 38.69 & 51.05 & 54.11 & 58.56 & 61.24 & 61.80 & 62.23 & 62.58 & 63.71 & 61.68 \\
DuReader & 17.51 & 17.55 & 16.75 & 16.80 & 17.42 & 17.51 & 17.36 & 16.88 & 17.48 & 16.48 \\
VCSUM & 5.36 & 9.57 & 11.61 & 12.47 & 13.25 & 13.52 & 12.77 & 13.64 & 12.67 & 12.60 \\
\midrule
Macro & 23.53 & 28.19 & 30.25 & 31.76 & 32.88 & 35.00 & 35.34 & 35.45 & 35.32 & 36.41 \\
\bottomrule
\end{tabular}%
}
\caption{Task-level scores at target retained fractions of the original context.
Macro is the unweighted mean of the same 12 tasks in each column.}
\label{tab:compression-ratios}
\end{table*}

\twocolumn
\makeatletter
\global\@colht=\textheight
\global\@colroom=\textheight
\global\vsize=\textheight
\makeatother
\section{Length-Stratified Compressor Profiling}
\label{sec:efficiency}
The profiling subset in Figure~\ref{fig:length_compression_cost}
is sampled with seed 42. The first and last length bins contain
five and four examples, respectively; all examples in the final
bin are from NarrativeQA.
The exact bin edges are 0, 2,048, 4,096, 8,192, 16,384,
and 32,768 document tokens, followed by an open-ended final bin.

Each method is measured in a separate process after loading its
compressor model. Examples are processed sequentially without
\texttt{empty\_cache()} between calls.
The wall clock around \texttt{compress()} is synchronized with CUDA
before and after the call. Peak memory is GPU allocated memory during
the call, including resident compressor weights, rather than reserved
memory. Within each bin, latency is averaged over successful
compression calls, and P95 is taken over their peak allocations.

Read uses Qwen2.5-0.5B with an all-layer attention probe;
Sentinel-0.5B uses the main detector on the same probe/SDPA path,
with \texttt{max\_seq\_len=32768} set explicitly.
All four series (Read, Sentinel-0.5B, LLMLingua-2, and CPC) were
measured under the same measurement protocol on the same machine
equipped with an NVIDIA RTX 5090 GPU.
The LLMLingua-2 and CPC measurements were collected in an earlier
profiling session on this same machine under the same measurement
protocol.
LLMLingua-2 uses its XLM-RoBERTa-large MeetingBank checkpoint with
default 512-token chunks and a final 2,000-token-budget repacking step.
CPC uses a Mistral-7B-Instruct backbone with CPC LoRA,
multi-chunk sentence scoring, and greedy sentence packing.

\clearpage
\twocolumn[{
\section{Head-Wise Attention Diagnostics}
\label{appendix:negative-heads}
\centering
\includegraphics[width=0.485\textwidth]{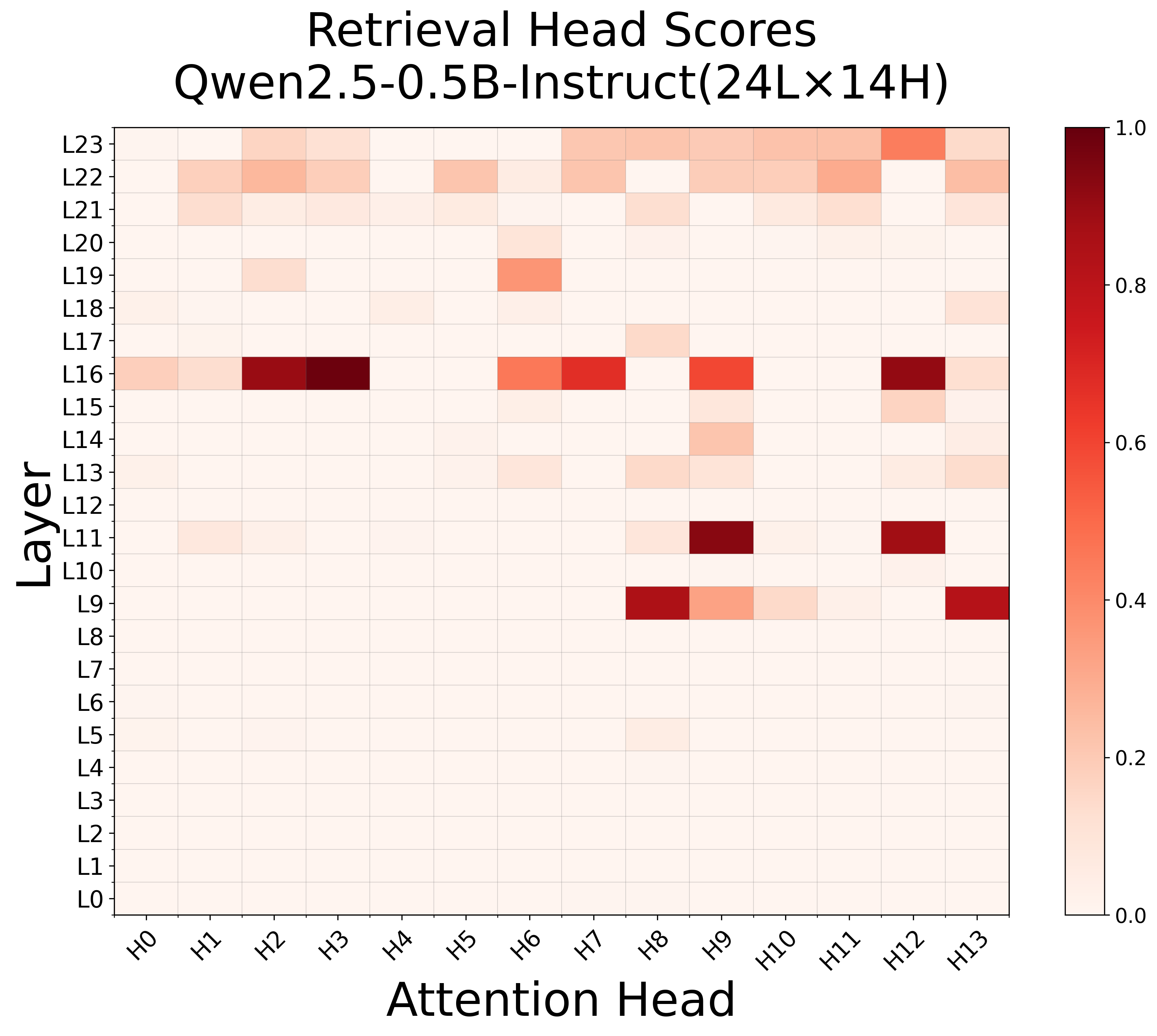}
\hfill
\includegraphics[width=0.485\textwidth]{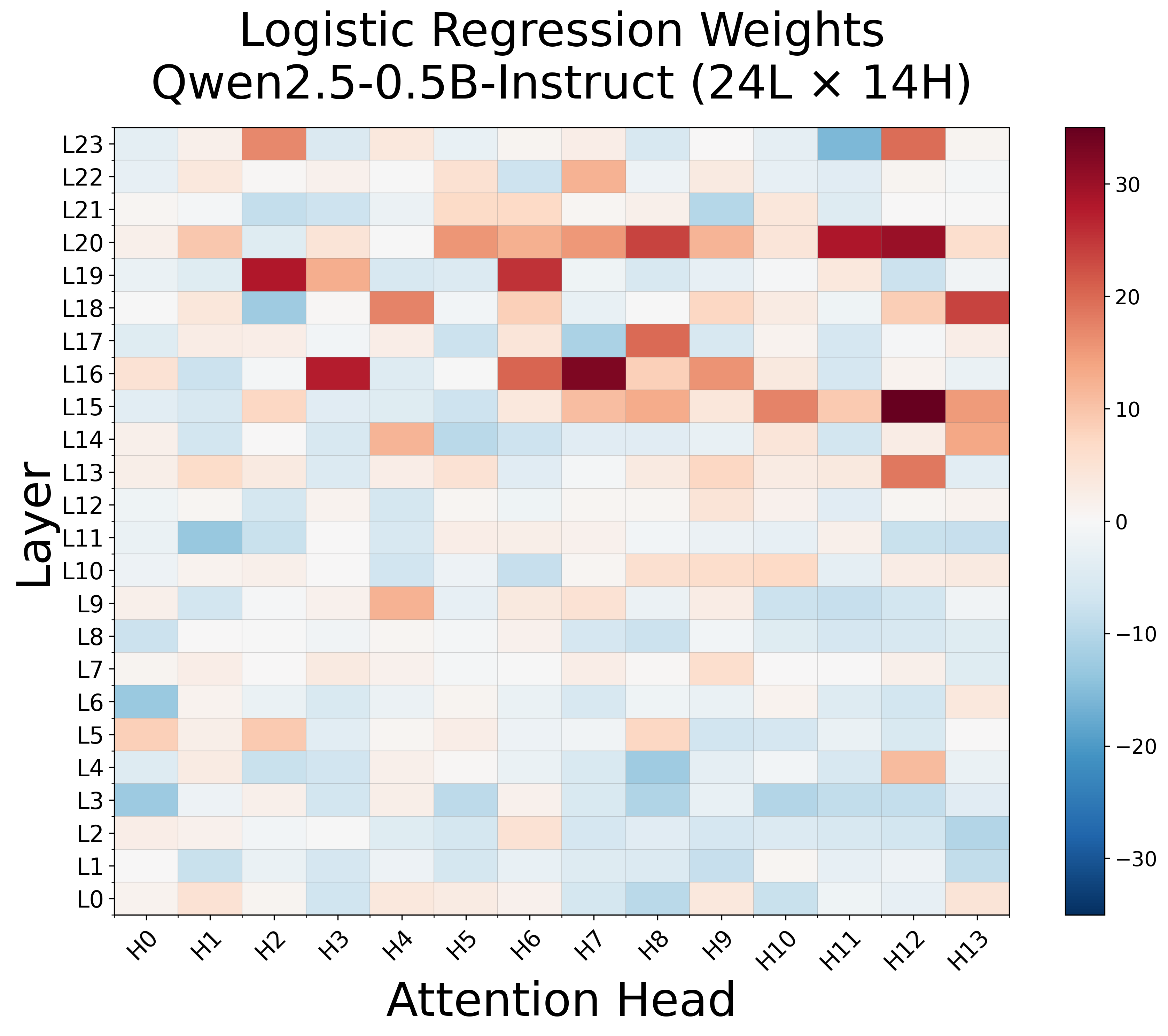}
\captionof{figure}{Head-wise patterns in Qwen2.5-0.5B-Instruct.
Left: retrieval-head scores. Right: Sentinel's signed probe coefficients.
Both panels index the same 24 layers and 14 heads per layer.}
\label{fig:head_diagnostics}
\par\vspace{1em}
}]

We examine selected negatively weighted heads of the Qwen2.5-0.5B
probe on 100 HotpotQA examples.
Tokens are grouped into sink tokens (including special tokens),
supporting-fact sentences, question tokens, and remaining context.
Table~\ref{tab:negative_heads} reports the mean attention mass
assigned to each group.
\newpage
\medskip
\noindent\begin{minipage}{\columnwidth}
    \centering
    \small
    \resizebox{\linewidth}{!}{
    \begin{tabular}{ccccccc}
        \toprule
        \textbf{Layer} & \textbf{Head} & \textbf{Probe Weight} & \textbf{Sink} & \textbf{Supporting} & \textbf{Question} & \textbf{Others} \\
        \midrule
        11 & 1  & -13.16 & 0.89 & 0.01 & 0.05 & 0.04 \\
        3  & 0  & -12.83 & 0.74 & 0.01 & 0.18 & 0.03 \\
        3  & 10 & -10.22 & 0.08 & 0.00 & 0.84 & 0.02 \\
        21 & 9  & -9.95  & 0.01 & 0.00 & 0.98 & 0.01 \\
        14 & 5  & -9.47  & 0.00 & 0.03 & 0.85 & 0.06 \\
        3  & 5  & -9.11  & 0.74 & 0.04 & 0.03 & 0.18 \\
        9  & 11 & -8.15  & 0.96 & 0.00 & 0.03 & 0.01 \\
        \bottomrule
    \end{tabular}
    }
\captionof{table}{
Mean attention mass by token group for heads with negative probe weights.
}
    \label{tab:negative_heads}
\end{minipage}
\medskip

\end{document}